\documentclass[runningheads]{llncs}
\usepackage{graphicx}
\usepackage{hyperref}
\usepackage{color}

\usepackage{times}
\usepackage{latexsym}
\usepackage{ulem}
\usepackage{graphicx}

\usepackage{url}

\usepackage[tracking=true]{microtype}

\usepackage{pifont}
\usepackage{amssymb}
\usepackage{booktabs}

\usepackage{fixltx2e}
\usepackage{arydshln}
\usepackage{cite}

\usepackage{booktabs}
\usepackage{tabularx}
\usepackage{multirow}
\usepackage{siunitx}

\newcolumntype{Y}{>{\centering\arraybackslash}X}

\begin{document}

%
\title{Contextualized Embeddings in Named-Entity Recognition: An Empirical Study on Generalization}
\titlerunning{Contextualized Embeddings in Named-Entity Recognition}
%

\author{Bruno Taill\'{e}\inst{1,2}\and
Vincent Guigue\inst{1} \and
Patrick Gallinari\inst{1,3}}
\authorrunning{B. Taill\'{e} et al.}

\institute{Sorbonne Universit\'{e}, CNRS, Laboratoire d'Informatique de Paris 6, LIP6 \and
BNP Paribas \and Criteo AI Lab, Paris\\
\email{\{bruno.taille, vincent.guigue, patrick.gallinari\}@lip6.fr}}


%
%
\maketitle              
\begin{abstract}
Contextualized embeddings use unsupervised language model pretraining to compute word representations depending on their context.
This is intuitively useful for generalization, especially in Named-Entity Recognition where it is crucial to detect mentions never seen during training.
However, standard English benchmarks overestimate the importance of lexical over contextual features because of an unrealistic lexical overlap between train and test mentions.
In this paper, we perform an empirical analysis of the generalization capabilities of state-of-the-art contextualized embeddings by separating mentions by novelty and with out-of-domain evaluation.
We show that they are particularly beneficial for unseen mentions detection, especially out-of-domain. For models trained on CoNLL03, language model contextualization leads to a +1.2\%  maximal relative micro-F1 score increase in-domain against +13\% out-of-domain on the WNUT dataset\footnotemark[1].
\keywords{NER \and Contextualized Embeddings \and Domain Adaptation}
\end{abstract}

\section{Introduction}
Named-Entity Recognition (NER) consists in detecting textual mentions of entities and classifying them into predefined types.
It is modeled as sequence labeling, the standard neural architecture of which is BiLSTM-CRF \cite{Huang2015BidirectionalTagging}. Recent improvements mainly stem from using new types of representations: learned character-level word embeddings \cite{Lample2016NeuralRecognition} and contextualized embeddings derived from a language model (LM) \cite{Peters2018DeepRepresentations, Akbik2018ContextualLabeling, Devlin2019BERT:Understanding}.

LM pretraining enables to obtain contextual word representations and reduce the dependency of neural networks on hand-labeled data specific to tasks or domains \cite{Howard2018UniversalClassification, Radford2018ImprovingPre-Training}. 
This contextualization ability can particularly benefit to NER domain adaptation which is often limited to training a network on source data and either feeding its predictions to a new classifier or finetuning it on target data \cite{Lee2018TransferNetworks, Rodriguez2018TransferClasses}.
All the more as classical NER models have been shown to poorly generalize to unseen mentions or domains \cite{Augenstein2017GeneralisationAnalysis}.


In this paper, we quantify the impact of ELMo \cite{Peters2018DeepRepresentations}, Flair \cite{Akbik2018ContextualLabeling} and BERT \cite{Devlin2019BERT:Understanding} representations on generalization to unseen mentions and new domains in NER. 
To better understand their effectiveness, we propose a set of experiments to distinguish the effect of unsupervised LM contextualization ($C_{LM}$) from task supervised contextualization ($C_{NER}$).
We show that the former mainly benefits unseen mentions detection, all the more out-of-domain where it is even more beneficial than the latter.
\footnotetext[1]{The code is available at \href{https://github.com/btaille/contener}{https://github.com/btaille/contener}}


\section{Lexical Overlap}
Neural NER models mainly rely on lexical features in the form of word embeddings, either learned at the character-level or not.
Yet, standard NER benchmarks present a large lexical overlap between mentions in the train set and dev / test sets which leads to a poor evaluation of generalization to unseen mentions as shown by Augenstein et al. \cite{Augenstein2017GeneralisationAnalysis}.
They separate seen from unseen mentions and evaluate out-of-domain to focus on generalization but only study models designed before 2011 and no longer in use.

We propose to use a similar setting to analyze the impact of state-of-the-art LM pretraining methods on generalization in NER.
We introduce a slightly more fine-grained novelty partition by separating unseen mentions in \textit{partial match} and \textit{new} categories. A mention is an \textit{exact match} (EM) if it appears in the exact same case-sensitive form in the train set, tagged with the same type.
It is a \textit{partial match} (PM) if at least one of its non stop words appears in a mention of same type. 
Every other mentions are \textit{new}.

We study lexical overlap in CoNLL03 \cite{Sang2003IntroductionTask} and OntoNotes \cite{Weischedel2013OntoNotesLDC2013T19}, the two main English NER datasets, as well as WNUT17 \cite{Derczynski2017ResultsRecognition} which is smaller, specific to user generated content (tweets, comments) and was designed without exact overlap. For out-of-domain evaluation, we train on CoNLL03 (news articles) and test on the larger and more diverse OntoNotes (see Table \ref{table:genres} for genres) and the very specific WNUT.
We remap OntoNotes and WNUT entity types to match CoNLL03's \footnotemark[1] and denote the obtained dataset with $^\ast$. 

\footnotetext[1]{We map LOC + GPE in OntoNotes to LOC in CoNLL03 and NORP + LANGUAGE to MISC. Other mappings are self-explanatory. We drop types that have no correspondence.}


\begin{table}[h]
\caption{Per type lexical overlap of test mention occurrences with respective train set in-domain and with CoNLL03 train set in the out-of-domain scenario. (EM / PM = \textit{exact / partial match})
}
\scriptsize
\begin{tabularx}{\textwidth}{@{}ll*{5}{Y}rYr*{5}{Y}rYr*{4}{Y}@{}}
\toprule
& &\multicolumn{5}{c}{CoNLL03} & & ON &  & \multicolumn{5}{c}{OntoNotes$^{\ast}$} & & WNUT & & \multicolumn{4}{c}{WNUT$^{\ast}$}\\
\cmidrule{3-7} \cmidrule{9-9} \cmidrule{11-15} \cmidrule{17-17} \cmidrule{19-22}
 & & LOC & MISC & ORG & PER & ALL & & ALL& & LOC & MISC & ORG & PER & ALL & &  ALL &  & LOC & ORG & PER & ALL\\ 
\midrule
\parbox[t]{3mm}{\multirow{3}{*}{\rotatebox[origin=c]{90}{\textbf{Self}}}} 
& EM & 82\% & 67\% & 54\% & 14\% & 52\% &
        & 67\% & 
       
        & 87\% & 93\% & 54\% & 49\% & 69\% &
         & - & 
         & - & - &- &- \\
        
& PM & 4\% & 11\% & 17\% & 43\% & 20\% & 
        & 24\% &
        
        & 6\% & 2\% & 32\% & 36\% & 20\% &
        & 12\% & 
        & 11\% & 5\% & 13\% & 12\% \\
        
& New  & 14\% & 22\% & 29\% & 43\% & 28\% &
        & 9\% & 
        
        & 7\% & 5\% & 14\% & 15\% & 11\% &
        & 88\% &
        & 89\% & 95\% & 87\% & 88\% \\
        
\midrule
\parbox[t]{3mm}{\multirow{3}{*}{\rotatebox[origin=c]{90}{\textbf{CoNLL}}}}  
        &  EM   &-  & - &  -& - &-  &
        &- & 
        
        & 70\% & 78\% & 18\% & 16\% & 42\%  &
        & - & 
        & 26\% & 8\% & 1\% & 7\%\\
        
& PM &-  & - &-  &  -& - & 
        & - &
        & 7\% & 10\% & 45\% & 46\% & 28\% &
        & - & 
        & 9\% & 15\% & 16\% & 14\% \\
        
& New  & -  & - & -  & - & - &
        & - & 
        & 23\% & 12\% & 38\% & 38\% & 30\%&
        & - &
        & 65\% & 77\% & 83\% & 78\%\\
        
\bottomrule
\end{tabularx}
\label{table:overlap}
\end{table}

As reported in Table \ref{table:overlap}, the two main benchmarks for English NER mainly evaluate performance on occurrences of mentions already seen during training, although they appear in different sentences. Such lexical overlap proportions are unrealistic in real-life where the model must process orders of magnitude more documents in the inference phase than it has been trained on, to amortize the annotation cost.
On the contrary, WNUT proposes a particularly challenging low-resource setting with no exact overlap.

Furthermore, the overlap depends on the entity types: Location and Miscellaneous are the most overlapping types, even out-of-domain, whereas Person and Organization present a more varied vocabulary, also more subject to evolve with time and domain.

\section{Word Representations}

\subsubsection{Word embeddings}
map each word to a single vector which results in a lexical representation. 
We take \textbf{GloVe 840B} embeddings \cite{Pennington2014GloVe:Representation} trained on Common Crawl as the pretrained word embeddings baseline and fine-tune them as done in related work.

\subsubsection{Character-level word embeddings}
are learned by a word-level neural network from character embeddings to incorporate orthographic and morphological features.
We reproduce the \textbf{Char-BiLSTM} from \cite{Lample2016NeuralRecognition}. 
It is trained jointly with the NER model and its outputs are concatenated to GloVe embeddings.
We also experiment with the Char-CNN layer from ELMo to isolate the effect of LM contextualization and denote it \textbf{ELMo[0]}.

\subsubsection{Contextualized word embeddings} take into account the context of a word in its representation, contrary to previous representations.
A LM is pretrained and used to predict the representation of a word given its context. 
\textbf{ELMo} \cite{Peters2018DeepRepresentations} uses a Char-CNN to obtain a context-independent word embedding and the concatenation of a forward and backward two-layer LSTM LM for contextualization.
These representations are summed with weights learned for each task as the LM is frozen after pretraining.
\textbf{BERT} \cite{Devlin2019BERT:Understanding} uses WordPiece subword embeddings \cite{Wu2016GooglesTranslation} and learns a representation 
modeling both left and right contexts by training a Transformer encoder \cite{Vaswani2017AttentionNeed} for Masked LM and next sentence prediction.
For a fairer comparison, we use the BERT\textsubscript{LARGE} feature-based approach where the LM is not fine-tuned and its last four hidden layers are concatenated.
\textbf{Flair} \cite{Akbik2018ContextualLabeling} uses a character-level LM for contextualization. 
As in ELMo, they train two opposite LSTM LMs, freeze them and concatenate the predicted states of the first and last characters of each word.
Flair and ELMo are pretrained on the 1 Billion Word Benchmark \cite{Chelba2013OneModeling} while BERT uses 
Book Corpus \cite{Zhu2015AligningBooks} and English Wikipedia.

\section{Experiments}

In order to compare the different embeddings, we feed them as input to a classifier.
We first use the state-of-the-art BiLSTM-CRF \cite{Huang2015BidirectionalTagging} with hidden size 100 in each direction and present in-domain results on all datasets in Table \ref{table:in-domain}.

We then report out-of-domain performance in Table \ref{table:results}. To better capture the intrinsic effect of LM contextualization, we introduce the Map-CRF baseline from \cite{Akbik2018ContextualLabeling} where the BiLSTM is replaced by a simple linear projection of each word embedding. We only consider domain adaptation from CoNLL03 to OntoNotes$^\ast$
and WNUT$^\ast$
assuming that labeled data is scarcer, less varied and more generic than target data in real use cases.


We use the IOBES tagging scheme for NER and no preprocessing. We fix a batch size of 64, a learning rate of 0.001 and a 0.5 dropout rate at the embedding layer and after the BiLSTM or linear projection.
The maximum number of epochs is set to 100 and we use early stopping with patience 5 on validation global micro-F1.
For each configuration, we use the best performing optimization method between SGD and Adam with $\beta_1=0.9$ and $\beta_2=0.999$. We report the mean and standard deviation of five runs.

\def\s{0.55}
\newcommand{\stddev}{\scalebox{\s}}{}
\def\dimspace{0.5em}

\begin{table}[h]
\caption{In-domain micro-F1 scores of the BiLSTM-CRF. We split mentions by novelty: \textit{exact match} (EM), \textit{partial match} (PM) and \textit{new}. Average of 5 runs, subscript denotes standard deviation. 
}

\centering
 \begin{tabularx}{\textwidth}{@{}lr@{\hspace{\dimspace}}*{4}{Y}r*{4}{Y}r*{3}{Y}@{}}

 \toprule
  &
 & \multicolumn{4}{c}{CoNLL03} & \hspace{\dimspace} & \multicolumn{4}{c}{OntoNotes$^{\ast}$} & \hspace{\dimspace} & \multicolumn{3}{c}{WNUT$^{\ast}$}\\
\cmidrule{3-6} \cmidrule{8-11} \cmidrule{13-15} 
Embedding  & Dim\hspace{\dimspace} & EM & PM & New & All & & EM & PM & New & All & & PM & New & All\\
\midrule

BERT                & 4096\hspace{\dimspace} & 95.7$_{\stddev{.1}}$ & 88.8$_{\stddev{.3}}$ & 82.2$_{\stddev{.3}}$ & 90.5$_{\stddev{.1}}$ &
                    & 96.9$_{\stddev{.2}}$ & 88.6$_{\stddev{.3}}$ & 81.1$_{\stddev{.5}}$ & \textbf{93.5}$_{\stddev{.2}}$ &
                    & 77.0$_{\stddev{4.6}}$ & 53.9$_{\stddev{.9}}$ & \textbf{57.0}$_{\stddev{1.0}}$\\

ELMo                & 1024\hspace{\dimspace} & 95.9$_{\stddev{.1}}$ & 89.2$_{\stddev{.5}}$ & 85.8$_{\stddev{.7}}$ & \textbf{91.8}$_{\stddev{.3  }}$ &
                    & 97.1$_{\stddev{.2}}$ & 88.0$_{\stddev{.2}}$ & 79.9$_{\stddev{.7}}$ & \textbf{93.4}$_{\stddev{.2}}$ &
                    & 67.7$_{\stddev{3.2}}$ & 49.5$_{\stddev{.9}}$ & 52.1$_{\stddev{1.0}}$\\

Flair               & 4096\hspace{\dimspace} & 95.4$_{\stddev{.1}}$ & 88.1$_{\stddev{.6}}$ & 83.5$_{\stddev{.5}}$ & 90.6$_{\stddev{.2}}$ &
                    & 96.7$_{\stddev{.1}}$ & 85.8$_{\stddev{.5}}$ & 75.0$_{\stddev{.6}}$ & 92.1$_{\stddev{.2}}$ &
                    & 64.9$_{\stddev{.7}}$ & 48.2$_{\stddev{2.0}}$ & 50.4$_{\stddev{1.8}}$\\

\cdashline{1-15}
ELMo[0]             & 1024\hspace{\dimspace}
                    & 95.8$_{\stddev{.1}}$ & 87.2$_{\stddev{.2}}$ & 83.5$_{\stddev{.4}}$ & 90.7$_{\stddev{.1}}$ &
                    & 96.9$_{\stddev{.1}}$ & 85.9$_{\stddev{.3}}$ & 75.5$_{\stddev{.6}}$ & 92.4$_{\stddev{.1}}$ &
                    & 72.8$_{\stddev{1.3}}$ & 45.4$_{\stddev{2.8}}$ & 49.1$_{\stddev{2.3}}$ \\
                    
GloVe + char & 350\hspace{\dimspace} 
                    & 95.3$_{\stddev{.3}}$ & 85.5$_{\stddev{.7}}$ & 83.1$_{\stddev{.7}}$ & 89.9$_{\stddev{.5}}$ &
                    & 96.3$_{\stddev{.1}}$ & 83.3$_{\stddev{.2}}$ & 69.9$_{\stddev{.6}}$ & 91.0$_{\stddev{.1}}$ &
                    & 63.2$_{\stddev{4.6}}$ & 33.4$_{\stddev{1.5}}$ & 38.0$_{\stddev{1.7}}$\\
                    
GloVe               & 300\hspace{\dimspace} 
                    & 95.1$_{\stddev{.4}}$ & 85.3$_{\stddev{.5}}$ & 81.1$_{\stddev{.5}}$ & 89.3$_{\stddev{.4}}$ &
                    & 96.2$_{\stddev{.2}}$ & 82.9$_{\stddev{.2}}$ & 63.8$_{\stddev{.5}}$ & 90.4$_{\stddev{.2}}$ &
                    & 59.1$_{\stddev{2.9}}$ & 28.1$_{\stddev{1.5}}$ & 32.9$_{\stddev{1.2}}$ \\
 \bottomrule
 \end{tabularx}

\label{table:in-domain}
\end{table}

 \begin{table}[h]
\caption{Micro-F1 scores of models trained on CoNLL03 and tested in-domain and out-of-domain on OntoNotes$^\ast$ and WNUT$^\ast$. 
Average of 5 runs, subscript denotes standard deviation.}

\centering

 \begin{tabularx}{\textwidth}{@{}l@{\hspace{0em}}lr*{4}{Y}r*{4}{Y}r*{4}{Y}@{}}
 \toprule


 &  & & \multicolumn{4}{c}{CoNLL03} & \hspace{.5em} & \multicolumn{4}{c}{OntoNotes$^\ast$} & \hspace{.5em} & \multicolumn{4}{c}{WNUT$^\ast$}\\
  \cmidrule{4-7} \cmidrule{9-12} \cmidrule{14-17} 
 & Emb & & EM & PM & New & All & & EM & PM & New & All & & EM & PM & New & All \\
\midrule

 \parbox[t]{4mm}{\multirow{6}{*}{\rotatebox[origin=c]{90}{\textbf{BiLSTM-CRF}}}} 
                      & BERT                   &    & 95.7$_{\stddev{.1}}$ & 88.8$_{\stddev{.3}}$ & 82.2$_{\stddev{.3}}$ & 90.5$_{\stddev{.1}}$ &
                                                    & 95.1$_{\stddev{.1}}$ & 82.9$_{\stddev{.5}}$ & 73.5$_{\stddev{.4}}$ & \textbf{85.0}$_{\stddev{.3}}$ & 
                                                    & 57.4$_{\stddev{1.0}}$ & 56.3$_{\stddev{1.2}}$ & 32.4$_{\stddev{.8}}$ & 37.6$_{\stddev{.8}}$\\
                                                    
                      & ELMo                   &    & 95.9$_{\stddev{.1}}$ & 89.2$_{\stddev{.5}}$ & 85.8$_{\stddev{.7}}$ & \textbf{91.8}$_{\stddev{.3}}$ &
                                                    & 94.3$_{\stddev{.1}}$ & 79.2$_{\stddev{.2}}$ & 72.4$_{\stddev{.4}}$ & 83.4$_{\stddev{.2}}$ &
                                                    & 55.8$_{\stddev{1.2}}$ & 52.7$_{\stddev{1.1}}$ & 36.5$_{\stddev{1.5}}$ & \textbf{41.0}$_{\stddev{1.2}}$\\
                                                    
                      & Flair                  &    & 95.4$_{\stddev{.1}}$ & 88.1$_{\stddev{.6}}$ & 83.5$_{\stddev{.5}}$ & 90.6$_{\stddev{.2}}$ &
                                                    & 94.0$_{\stddev{.3}}$ & 76.1$_{\stddev{1.1}}$ & 62.1$_{\stddev{.5}}$ & 79.0$_{\stddev{.5}}$ &
                                                    & 56.2$_{\stddev{2.2}}$ & 49.4$_{\stddev{3.4}}$ & 29.1$_{\stddev{3.3}}$ & 34.9$_{\stddev{2.9}}$\\
\cdashline{2-17}
                      & ELMo[0] &    & 95.8$_{\stddev{.1}}$ & 87.2$_{\stddev{.2}}$ & 83.5$_{\stddev{.4}}$ & 90.7$_{\stddev{.1}}$ &
                                                    & 93.6$_{\stddev{.1}}$ & 76.8$_{\stddev{.6}}$ & 66.1$_{\stddev{.3}}$ & 80.5$_{\stddev{.2}}$ &
                                                    & 52.3$_{\stddev{1.2}}$ & 50.8$_{\stddev{1.5}}$ & 32.6$_{\stddev{2.2}}$ & 37.6$_{\stddev{1.8}}$\\
                                                    
                      & G + char              &     & 95.3$_{\stddev{.3}}$ & 85.5$_{\stddev{.7}}$ & 83.1$_{\stddev{.7}}$ & 89.9$_{\stddev{.5}}$ &
                                                    & 93.9$_{\stddev{.2}}$ & 73.9$_{\stddev{1.1}}$ & 60.4$_{\stddev{.7}}$ & 77.9$_{\stddev{.5}}$ &
                                                    & 55.9$_{\stddev{.8}}$ & 46.8$_{\stddev{1.8}}$ & 19.6$_{\stddev{1.6}}$ & 27.2$_{\stddev{1.3}}$\\
                                                    
                      & GloVe                  &    & 95.1$_{\stddev{.4}}$ & 85.3$_{\stddev{.5}}$ & 81.1$_{\stddev{.5}}$ & 89.3$_{\stddev{.4}}$ &
                                                    & 93.7$_{\stddev{.2}}$ & 73.0$_{\stddev{1.2}}$ & 57.4$_{\stddev{1.8}}$ & 76.9$_{\stddev{.9}}$ &
                                                    & 53.9$_{\stddev{1.2}}$ & 46.3$_{\stddev{1.5}}$ & 13.3$_{\stddev{1.4}}$ & 27.1$_{\stddev{1.0}}$\\
\cmidrule{1-17}
\parbox[t]{4mm}{\multirow{6}{*}{\rotatebox[origin=c]{90}{\textbf{Map-CRF}}}}
                      & BERT                   &    & 93.2$_{\stddev{.3}}$ & 85.8$_{\stddev{.4}}$ & 73.7$_{\stddev{.8}}$ & 86.2$_{\stddev{.4}}$ &
                                                    & 93.5$_{\stddev{.2}}$ & 77.8$_{\stddev{.5}}$ & 67.8$_{\stddev{.9}}$ & 80.9$_{\stddev{.4}}$ &
                                                    & 57.4$_{\stddev{.3}}$ & 53.5$_{\stddev{2.6}}$ & 33.9$_{\stddev{.6}}$ & 38.4$_{\stddev{.4}}$\\
                                                    
                      & ELMo                   &    & 93.7$_{\stddev{.2}}$ & 87.2$_{\stddev{.6}}$ & 80.1$_{\stddev{.3}}$ & \textbf{88.7}$_{\stddev{.2}}$ &
                                                    & 93.6$_{\stddev{.1}}$ & 79.1$_{\stddev{.5}}$ & 69.5$_{\stddev{.4}}$ & \textbf{82.2}$_{\stddev{.3}}$ &
                                                    & 61.1$_{\stddev{.7}}$ & 53.0$_{\stddev{.9}}$ & 37.5$_{\stddev{.7}}$ & \textbf{42.4}$_{\stddev{.6}}$\\

                      & Flair                   &   & 94.3$_{\stddev{.1}}$ & 85.1$_{\stddev{.3}}$ & 78.6$_{\stddev{.3}}$ & 88.1$_{\stddev{.03}}$ &
                                                    & 93.2$_{\stddev{.1}}$ & 74.0$_{\stddev{.3}}$ & 59.6$_{\stddev{.2}}$ & 77.5$_{\stddev{.2}}$ &
                                                    & 52.5$_{\stddev{1.2}}$ & 50.6$_{\stddev{.4}}$ & 28.8$_{\stddev{.5}}$ & 33.7$_{\stddev{.5}}$\\
\cdashline{2-17}
                      & ELMo[0]                 &   & 92.2$_{\stddev{.3}}$ & 80.5$_{\stddev{1.0}}$ & 68.6$_{\stddev{.4}}$ & 83.4$_{\stddev{.4}}$ &
                                                    & 91.6$_{\stddev{.4}}$ & 69.6$_{\stddev{1.0}}$ & 56.8$_{\stddev{1.5}}$ & 75.0$_{\stddev{1.0}}$ &
                                                    & 51.9$_{\stddev{1.1}}$ & 42.6$_{\stddev{.9}}$ & 32.4$_{\stddev{.3}}$ & 35.8$_{\stddev{.4}}$\\

                      & G + char                &   & 93.1$_{\stddev{.3}}$ & 80.7$_{\stddev{.9}}$ & 69.8$_{\stddev{.7}}$ & 84.4$_{\stddev{.4}}$ &
                                                    & 91.8$_{\stddev{.3}}$ & 69.3$_{\stddev{.3}}$ & 55.6$_{\stddev{1.1}}$ & 74.8$_{\stddev{.5}}$ & 
                                                    & 50.6$_{\stddev{.9}}$ & 42.5$_{\stddev{1.4}}$ & 20.6$_{\stddev{2.8}}$ & 28.7$_{\stddev{2.5}}$\\
                                                    
                      & GloVe                   &   & 92.2$_{\stddev{.1}}$ & 77.0$_{\stddev{.4}}$ & 61.7$_{\stddev{.3}}$ & 81.5$_{\stddev{.05}}$ &
                                                    & 89.6$_{\stddev{.3}}$ & 62.8$_{\stddev{.6}}$ & 38.5$_{\stddev{.4}}$ & 68.1$_{\stddev{.4}}$ & 
                                                    & 46.8$_{\stddev{.8}}$ & 41.3$_{\stddev{.5}}$ & 3.2$_{\stddev{.2}}$ & 18.9$_{\stddev{.7}}$\\
 \bottomrule
 \end{tabularx}

\label{table:results}
\end{table}


\subsection{General Observations} \label{general}

\subsubsection{ELMo, BERT and Flair}
Drawing conclusions from the comparison of ELMo, BERT and Flair is difficult because there is no clear hierarchy accross datasets and they differ in dimensions, tokenization, contextualization levels and pretraining corpora.
However, although BERT is particularly effective on the WNUT dataset in-domain, probably due to its subword tokenization, ELMo yields the most stable results in and out-of-domain.

Furthermore, Flair globally underperforms ELMo and BERT, particularly for unseen mentions and out-of-domain. 
This suggests that LM pretraining at a lexical level (word or subword) is more robust for generalization than at a character level.
In fact, Flair only beats the non contextual ELMo[0] baseline with Map-CRF which indicates that character-level contextualization is less beneficial than word-level contextualization with character-level representations.


\subsubsection{ELMo[0] vs GloVe+char}

Overall, ELMo[0] outperforms the GloVe+char baseline, particularly
on unseen mentions, out-of-domain and on WNUT$^\ast$.
The main difference is the incorporation of morphological features: in ELMo[0] they are learned jointly with the LM on a huge dataset whereas the char-BiLSTM is only trained on the source NER training set. Yet, morphology is crucial to represent words never encountered during pretraining and in WNUT$^\ast$ around 20\% of words in test mentions are out of GloVe vocabulary against 5\% in CoNLL03 and 3\% in OntoNotes$^\ast$. This explains the poor performance of GloVe baselines on WNUT$^\ast$, all the more out-of-domain, and why a model trained on CoNLL03 with ELMo outperforms one trained on WNUT$^\ast$ with GloVe+char.
Thus, ELMo's improvement over previous state-of-the-art does not
only stem from contextualization but also an effective non-contextual word representation.




\subsubsection{Seen Mentions Bias}
In every configuration, 
$F1_{exact} > F1_{partial} > F1_{new}$
with more than 10 points difference.
This gap is wider out-of-domain where the context differs more from training data than in-domain.
NER models thus poorly generalize to unseen mentions, and datasets with high lexical overlap only encourage this behavior.
However, this generalization gap is reduced by two types of contextualization described hereafter.

\subsection{LM and NER Contextualizations} \label{analysis}

The ELMo[0] and Map-CRF baselines enable to strictly distinguish contextualization due to LM pretraining ($C_{LM}$: ELMo[0] to ELMo) from task supervised contextualization induced by the BiLSTM network ($C_{NER}$: Map to BiLSTM).
In both cases, a BiLSTM incorporates syntactic information which improves generalization to unseen mentions for which context is decisive, as shown in Table \ref{table:results}.


\subsubsection{Comparison}

However, because ${C_{NER}}$ is specific to the source dataset, it is more effective in-domain whereas $C_{LM}$ is particularly helpful out-of-domain. 
In the latter setting, the benefits from $C_{LM}$ even surpass those from ${C_{NER}}$, specifically on domains further from source data such as web text in OntoNotes$^\ast$ (see Table \ref{table:genres}) or WNUT$^\ast$.
This is again explained by the difference in quantity and quality of the corpora on which these contextualizations are learned.
The much larger and more generic unlabeled corpora on which LM are pretrained lead to contextual representations more robust to domain adaptation than ${C_{NER}}$ learned on a small and specific NER corpus.

Similar behaviors can be observed when comparing BERT and Flair to the GloVe baselines, although we cannot separate the effects of representation and contextualization.
\subsubsection{Complementarity}

Both in-domain and out-of-domain on OntoNotes$^\ast$, the two types of contextualization transfer complementary syntactic features leading to the best configuration. However,  in the most difficult case of zero-shot domain adaptation from CoNLL03 to WNUT$^\ast$, ${C_{NER}}$ is detrimental with ELMo and BERT. This is probably due to the specificity of the target domain, excessively different from source data.

\def\s{0.7}
\newcommand{\stddevtwo}{\scalebox{\s}}{}

\begin{table}[h!]
\centering
\caption{Per-genre micro-F1 scores of the BiLSTM-CRF model trained on CoNLL03 and tested on OntoNotes$^\ast$ (broadcast conversation, broadcast news, news wire, magazine, telephone conversation and web text). $C_{LM}$ mostly benefits genres furthest from the news source domain. }

 \begin{tabularx}{0.75\textwidth}{@{}l*{7}{Y}@{}}
 \toprule
  & bc & bn & nw & mz & tc & wb & All\\
  \midrule
 BERT           & 87.2$_{\stddevtwo{.5}}$ & 88.4$_{\stddevtwo{.4}}$ & 84.7$_{\stddevtwo{.2}}$ & 82.4$_{\stddevtwo{1.2}}$ & 84.5$_{\stddevtwo{1.1}}$ & 79.5$_{\stddevtwo{1.0}}$ & \textbf{85.0}$_{\stddevtwo{.3}}$ \\
 ELMo           & 85.0$_{\stddevtwo{.6}}$ & 88.6$_{\stddevtwo{.3}}$ & 82.9$_{\stddevtwo{.3}}$ & 78.1$_{\stddevtwo{.7}}$ & 84.0$_{\stddevtwo{.8}}$ & 79.9$_{\stddevtwo{.5}}$ & 83.4$_{\stddevtwo{.2}}$ \\
 Flair          & 78.0$_{\stddevtwo{1.1}}$ & 86.5$_{\stddevtwo{.4}}$ & 80.4$_{\stddevtwo{.6}}$ & 71.1$_{\stddevtwo{.4}}$ & 73.5$_{\stddevtwo{1.8}}$ & 72.1$_{\stddevtwo{.8}}$ & 79.0$_{\stddevtwo{.5}}$ \\
 \cdashline{1-8}
 ELMo[0]        & 82.6$_{\stddevtwo{.5}}$ & 88.0$_{\stddevtwo{.3}}$ & 79.6$_{\stddevtwo{.5}}$ & 73.4$_{\stddevtwo{.6}}$ & 79.2$_{\stddevtwo{1.2}}$ & 75.1$_{\stddevtwo{.3}}$ & 80.5$_{\stddevtwo{.2}}$ \\
 GloVe + char \hspace{2em}& 80.4$_{\stddevtwo{.8}}$ & 86.3$_{\stddevtwo{.4}}$ & 77.0$_{\stddevtwo{1.0}}$ & 70.7$_{\stddevtwo{.4}}$ & 79.7$_{\stddevtwo{1.8}}$ & 69.2$_{\stddevtwo{.8}}$ & 77.9$_{\stddevtwo{.5}}$ \\
 \bottomrule
 \end{tabularx}

\label{table:genres}
\end{table}

\section{Related Work}
Augenstein et al. \cite{Augenstein2017GeneralisationAnalysis} perform a quantitative study of two CRF-based models and a CNN with classical word embeddings \cite{Collobert2011NaturalScratch} over seven NER datasets including CoNLL03 and OntoNotes.
They separate performance on seen (\textit{exact match}) and unseen mentions and show a drop in F1 on unseen mentions and out-of-domain.
Although comprehensive in experiments, this analysis is limited to models dating back from 2005 to 2011.
We use a similar experimental setting to draw new insights on state-of-the art architectures and word representations.
We limit to the two main English NER benchmarks as well as WNUT which was specifically designed to tackle this generalization problem in the Twitter domain. These three datasets cover all the domains studied in \cite{Augenstein2017GeneralisationAnalysis}.

Moosavi and Strube raise a similar lexical overlap issue in Coreference Resolution on the CoNLL2012 dataset. 
They first
 show that for out-of-domain evaluation the performance gap between Deep Learning models and a rule-based system fades away \cite{Moosavi2017LexicalCaution}. 
They then add linguistic features (such as gender, NER, POS...) to improve out-of-domain generalization \cite{Moosavi2018UsingResolvers}. 
 Nevertheless, such features are obtained using models in turn based on lexical features and at least for NER the same lexical overlap issue arises.

Finally, Pires et al. \cite{Pires2019HowBERT} concurrently evaluate the cross-lingual generalization capability of Multilingual BERT for NER and POS tagging. Our work on monolingual generalization to unseen mentions and domains naturally complements this study.

\section{Conclusion}
NER benchmarks are biased towards seen mentions, at the opposite of real-life applications.
Hence the necessity to disentangle performance on seen and unseen mentions and test out-of-domain.
In such setting, we show that contextualization from LM pretraining is particularly beneficial for generalization to unseen mentions, all the more out-of-domain where it surpasses supervised contextualization.

Despite this improvement, unseen mentions detection remains challenging and further work could explore attention or regularization mechanisms to better incorporate context and improve generalization. Furthermore, we can investigate how to best incorporate target data to improve this LM pretraining zero-shot domain adaptation baseline.




\bibliographystyle{splncs04}
\bibliography{mendeley}

\end{document}